\documentclass[11pt]{article}
\usepackage[a4paper,margin=1in]{geometry}
\usepackage[T1]{fontenc}
\usepackage[utf8]{inputenc}
\usepackage{lmodern}
\usepackage{amsmath,amssymb,bm}
\usepackage{graphicx}
\usepackage{booktabs}
\usepackage{array,tabularx}
\usepackage{tabu}
\usepackage{multicol}
\usepackage{algorithm}
\usepackage{algorithmic}
\usepackage{url}
\usepackage[hidelinks]{hyperref}
\usepackage{microtype}
\usepackage{caption}
\usepackage{float}

\title{A Hybrid PEM--GP Framework for Uncertainty-Aware System Identification of Quadcopters}
\author{Abdallah Ghoul$^{1,*}$ \and Ismail Khalil Bousserhane$^{1}$ \and Kadri Boufeldja$^{1}$}
\date{}

\begin{document}
\maketitle

\begin{center}
\small
$^{1}$Electrical Engineering Department, University of Bechar, Bechar 08000, Algeria\\
$^{*}$Corresponding author: \texttt{abdallah.ghoul@univ-bechar.dz}\\[0.5em]
\textbf{Journal reference:} International Journal of Control, Automation, and Systems, Vol. 24, No. 8, pp. 2047--2057, 2026.\\
\textbf{DOI:} \href{https://doi.org/10.1007/s12555-026-00123-5}{10.1007/s12555-026-00123-5}
\end{center}

\begin{abstract}
Accurate dynamic models play a central role in achieving reliable control of quadcopters. Classical system identification methods remain widely used, mainly because of their interpretability. However, they often fail to capture important nonlinear effects, especially in small-scale aerial platforms where such effects become more pronounced.

Data-driven approaches offer a different perspective. They can represent complex nonlinear dynamics more effectively, but this comes at the cost of reduced interpretability and the absence of well-calibrated uncertainty estimates.

In this work, we propose a framework that combines physics-based modeling with data-driven learning, while explicitly accounting for uncertainty. A physics-based model is first identified using the Prediction Error Method (PEM), which captures the main structure of the system. The remaining dynamics are then modeled using a Gaussian Process (GP), allowing the residual behavior to be learned directly from data. This separation makes it possible to distinguish between known physical effects and unmodeled dynamics.

The proposed framework is validated on a Duckiedrone-like experimental setup. The results show that the PEM--GP model achieves prediction accuracy comparable to that of a Long Short-Term Memory (LSTM) network, while additionally providing calibrated uncertainty estimates. This combination improves model reliability and supports uncertainty-aware decision-making.
\end{abstract}


\section{INTRODUCTION}
	
	Unmanned aerial vehicles (UAVs)—and especially quadcopters—now serve research, inspection, agriculture, and last-mile delivery \cite{smith2020applications, johnson2021future}. Their value depends on precise, stable flight. That, in turn, depends on a dynamic model that predicts the vehicle response under control inputs and disturbances with high fidelity \cite{brescianini2018nonlinear}.
	System identification provides that model. It estimates parameters and structures directly from data \cite{ljung1999system}. For quadcopters this includes inertias, thrust and drag coefficients, aerodynamic derivatives, and trajectory-level motion primitives \cite{mueller2015system, mueller2015computationally}. The quality of the identified model controls what follows: model-based control, high-fidelity simulation, and state estimation \cite{mueller2015system}. If the model is weak, everything downstream suffers.
	Small platforms such as Duckiedrone make this hard \cite{duckietown2023drone}. They are light, underactuated, and sensitive to effects that larger vehicles mostly ignore. Examples include PWM-to-thrust nonlinearity and voltage sag, blade flapping and ground effect, and strong cross-axis coupling \cite{faessler2016modeling, bangura2016aerodynamics}. Low-cost MEMS IMUs add noise, bias, and scale errors. Data look messy; dynamics hide behind sensor artifacts.
	Classical identification remains a useful baseline. PEM and frequency-domain analysis work well for linear, time-invariant systems and yield interpretable models \cite{ljung1999system}. On small quadcopters their scope is narrow. Linearization around hover helps, but aggressive flight and aerodynamic nonlinearities break assumptions. Treating roll, pitch, and yaw as decoupled also fails once manoeuvres couple axes. The models hold only near a single operating point \cite{schiano2018comparison}.
	Data-driven methods push farther. LSTM networks learn nonlinear temporal dependencies and often predict better than classical models \cite{gamboa2017deep}. They are black boxes, though. They do not explain their outputs and, critically, they do not report uncertainty. That is a problem in safety-critical robotics \cite{deisenroth2013survey}. GP offer a probabilistic alternative with calibrated uncertainty \cite{rasmussen2006gaussian}. A full GP over the entire dynamics can be data-hungry and may ignore known physics.
	We take a hybrid path. A simplified PEM-based physics model captures dominant, structured dynamics. A GP then learns the residual—the part the physics misses. The physics keeps interpretability and acts as a strong prior. The GP adds flexibility and delivers uncertainty. Together they cover the space that neither covers alone.
	
	Recent advances in learning-based system identification have explored hybrid modeling strategies that combine physics-based models with machine learning techniques. Several studies augment simplified UAV dynamic models using learning algorithms to capture aerodynamic disturbances and nonlinear effects that are difficult to model analytically. For example, Becker et al. demonstrated that learning-based aerodynamic compensation can improve quadrotor prediction accuracy during aggressive maneuvers \cite{becker2022learning}. Similarly, Faessler et al. showed that aerodynamic effects such as blade flapping and drag forces play a significant role in quadrotor dynamics and must be properly modeled to obtain accurate system identification results \cite{faessler2016modeling}.

Gaussian Process regression has also gained increasing attention in robotics because it provides probabilistic predictions and calibrated uncertainty estimates. These properties are particularly important for safety-critical systems such as UAVs where controllers must account for model confidence. The theoretical foundations of Gaussian Process modeling are presented in Rasmussen and Williams \cite{rasmussen2006gaussian}, while Deisenroth et al. demonstrated the effectiveness of probabilistic dynamics models for robotics learning and control tasks \cite{deisenroth2013survey}.

Hybrid identification approaches aim to combine the strengths of physical modeling and data-driven learning. Physics-based models preserve interpretability and incorporate known structural properties of the system, whereas learning algorithms can capture residual nonlinear dynamics that remain unmodeled. However, many existing approaches either rely entirely on black-box models such as neural networks \cite{hochreiter1997lstm} or use physics models only as initialization for data-driven learning.

Explicit frameworks that combine Prediction Error Method identification with Gaussian Process residual learning for quadrotor dynamics remain relatively limited in the UAV identification literature. In particular, the use of a PEM-based parametric model as a structured baseline together with a GP residual learner provides a principled way to distribute modeling effort between known physical dynamics and unmodeled nonlinear effects. In this work, we investigate such a hybrid identification strategy where a PEM-based parametric model captures the dominant structured dynamics, while a Gaussian Process learns the residual nonlinear behavior that the physics model cannot represent.
	
	The principal contributions of this work are threefold:
	
	We introduce a novel, structured system identification methodology that hybridizes a classical PEM model with a GP residual learner for small-scale quadcopters.
	
	We demonstrate that this hybrid PEM-GP framework not only achieves high predictive accuracy, comparable to a black-box LSTM, but also provides explicit, quantitative uncertainty estimates—a feature absent in both PEM and LSTM models.
	
	We provide a comprehensive experimental evaluation using the Duckiedrone model, validating the performance of our approach against standalone PEM, GP, and LSTM baselines across a range of excitation trajectories.
	
	The remainder of this paper is organized as follows: Section 2 reviews related work in classical, data-driven, and hybrid system identification. Section 3 details our methodology, including the problem formulation and the proposed hybrid framework. Section 4 describes the experimental setup. Section 5 presents and discusses the results, and Section 6 offers concluding remarks and directions for future work.
	
	\section{Related Work}
	\label{sec:related_work}
	Research on system identification for robots spans three lines:
	classical parametric methods, data-driven non-parametric methods, and Grey-box hybrids.
	This section reviews representative work in each line with an emphasis on UAVs.
	\subsection{Classical System Identification Methods}
	Classical identification offers a rigorous foundation.
	Ljung formalized the PEM and its guarantees for linear, time-invariant models \cite{ljung1999system}.
	For UAVs, Mueller et al. demonstrated PEM on a small quadcopter and showed its utility for controller design \cite{mueller2015system}. In related work, efficient motion primitives for quadrotor trajectory generation have also been proposed, highlighting the importance of accurate dynamic models for real-time applications \cite{mueller2015computationally}.
	Frequency-domain techniques remain attractive in aerospace for their noise rejection and clear gain-phase interpretation \cite{schiano2018comparison}.
	
	Limits appear on small, agile platforms.
	The linearity assumption leaves out key effects \cite{brescianini2018nonlinear}.
	Axis-by-axis identification ignores cross-couplings that the aerodynamics literature documents \cite{faessler2016modeling, bangura2016aerodynamics}.
	In practice, these models work near hover and degrade under coupled, aggressive motion.
	They are interpretable and fast, but their valid region is narrow.
	
	\subsection{Data-Driven and Machine Learning Approaches}
	Machine learning addresses those limits by learning dynamics directly from data.
	Neural networks are universal approximators; see Gamboa for a concise overview of deep models for sequences \cite{gamboa2017deep}.
	Recurrent architectures—and LSTMs in particular—capture temporal dependencies and often outperform linear baselines on complex UAV trajectories \cite{hochreiter1997lstm}.
	
	Gaussian Processes provide a Bayesian alternative.
	Rasmussen and Williams summarize the theory and the role of kernels and uncertainty \cite{rasmussen2006gaussian}.
	In robotics, GP are valued for calibrated prediction variance, which supports safe decision-making and robust control \cite{deisenroth2013survey}.
	
	Despite their strong modeling capabilities, these approaches present several limitations. Deep neural networks are typically considered black-box models, offering limited physical interpretability and reduced transparency in decision-making processes \cite{rudin2019stop}. In addition, GP models may suffer from scalability issues when applied to large datasets and can become data-inefficient when used to learn full system dynamics without incorporating prior physical knowledge.
	
	\subsection{Grey-Box and Hybrid Modeling Techniques}
	Grey-box methods combine physics with learning.
	A physics model captures dominant structure, then a learning module explains residual dynamics.
	
	Prior work explores several patterns:
	Neural networks augment specific unmodeled terms such as aerodynamic wrenches \cite{becker2022learning}.
	Simple physics models paired with GP have been used in manipulation and related tasks \cite{ko2007gp}.
	However, for full quadcopter identification, systematic frameworks combining a PEM-based baseline with GP residual learning remain relatively scarce in the literature.
	
	Our approach fills that gap.
	We first identify a PEM-based white-box model for the core dynamics.
	We then train a GP on the residual error to capture nonlinearities and cross-axis couplings the PEM misses.
	The split focuses learning capacity where it matters while keeping a physically meaningful core.
	The result is a practical balance: high accuracy, explicit uncertainty, and retained interpretability.
	
	Although hybrid physics–learning approaches have been explored in robotics and system identification, many existing studies either rely entirely on data-driven models or augment physics-based models using machine learning components \cite{deisenroth2015pilco,faessler2016modeling}. For instance, Gaussian Process models have been widely used for learning system dynamics and uncertainty-aware control in robotics \cite{rasmussen2006gaussian,deisenroth2013survey}. 

In contrast, the proposed framework combines Prediction Error Method (PEM) identification with Gaussian Process residual learning in a structured identification pipeline specifically designed for quadrotor dynamics. In this approach, the PEM model captures the dominant physics-based behavior of the system, while the GP is trained only on the residual nonlinear dynamics that remain unexplained by the parametric model. This separation allows the learning component to focus on a simpler modeling task while preserving the interpretability of the physical model parameters.
	
	\section{Methodology}
	\label{sec:methodology}
	
	This section outlines the theoretical foundation and the proposed framework for quadcopter system identification. We begin with the problem formulation and a description of the standard quadcopter dynamics, followed by an overview of the baseline identification methods. Finally, we present our novel hybrid PEM-GP methodology in detail.
	\subsection{Problem Formulation and Quadcopter Dynamics}
	
	We pose quadcopter identification as supervised sequence modeling.
	Given an input–output datasets: 
	\begin{equation}
		\mathcal{D} = \{u(t), y(t)\}_{t=1}^{N} 
	\end{equation}
	
	With motor commands $u(t) \in \mathbb{R}^{4}$ (PWM values, squared rotor speeds) and measured outputs $y(t) \in \mathbb{R}^{m}$ (angular rates, linear accelerations), the goal is to learn a predictive model $\mathcal{M}$ that maps inputs to outputs:
	
	\begin{equation}
		\hat{y}(t)= \mathcal{M}(u(t),u(1:t-1),y(1:t-1)),
	\end{equation}
	
	such that $\hat{y}(t)$ accurately reproduces the system response for the given $u(t)$. Identification is cast as minimizing a prediction-error objective over $\mathcal{D}$.
	
	The body-frame Newton–Euler equations are:
	\begin{equation}
		\label{eq:newton_euler}
		\begin{aligned}
			m\dot{\mathbf v} + \boldsymbol{\omega}\times(m\mathbf v) &= \mathbf F,\\
			\mathbf{J}\dot{\boldsymbol{\omega}} + \boldsymbol{\omega}\times(\mathbf J\boldsymbol{\omega}) &= \boldsymbol{\tau},
		\end{aligned}
	\end{equation}
	where \( m \) is the mass, \( \mathbf{v} = [u, v, w]^T \) is the linear velocity vector, \( \boldsymbol{\omega} = [p, q, r]^T \) is the angular velocity vector, \( \mathbf{J} \) is the inertia matrix, \( \mathbf{F} \) is the total force vector, and \( \boldsymbol{\tau} = [\tau_{\phi}, \tau_{\theta}, \tau_{\psi}]^T \) is the total torque vector.
	
	The total force $\mathbf{F}$ and torque $\boldsymbol{\tau}$ are:
	
	\begin{equation}
		\label{eq:forces_torques}
		\begin{aligned}
			\mathbf{F} &= \begin{bmatrix} 0 \\ 0 \\ \sum_{i=1}^{4} F_i \end{bmatrix} + \mathbf{R}^T \begin{bmatrix} 0 \\ 0 \\ -mg \end{bmatrix} + \mathbf{F}_{a}, \\ \\
			\boldsymbol{\tau} &= \begin{bmatrix}
				l (F_2 - F_4) \\
				l (F_3 - F_1) \\
				\kappa (F_1 - F_2 + F_3 - F_4)
			\end{bmatrix} + \boldsymbol{\tau}_{a},
		\end{aligned}
	\end{equation}
	
	with $F_i=c_t \omega_i^2$ the $i$-th rotor thrust, $l$ the arm length, $\kappa$ the yaw torque coefficient, $\mathbf{R}$ the rotation matrix (world to body), $g$ gravity, and $\boldsymbol{F}_{a}$, $\boldsymbol{\tau}_{a}$ aerodynamic contributions not captured by the nominal model.
	
	\subsection{Baseline Identification Methods}
	
	\subsubsection{Prediction Error Method (PEM)}
	In this work we use PEM \cite{ljung1999system} as the classical baseline. 
	
	Given a parametric model $\mathcal{M}(\theta)$ and data $\{u(t),y(t)\}_{t=1}^N$ , PEM estimates $\theta$ by minimizing the one–step prediction error:
	\begin{equation}
		\label{eq:pem}
		\hat{\theta} = \arg \min_{\theta} J_N(\theta), \ \ J_N(\theta)=\frac{1}{2} \sum_{t=1}^{N} \lVert e(t,\theta) \rVert^2 ,
	\end{equation}
	
	where $e(t,\theta)=y(t)-\hat{y}(t|t-1;\theta)$ is the conditional one-step-ahead predictor.
	
	We consider discrete-time , linear state-space models(innovation form):
	
	\begin{equation}
		\begin{aligned}
			x_{k+1}&=Ax_k+ Bu_k+Ke_k,\\
			y_k&=Cx_k+Du_k+e_k,
		\end{aligned}
	\end{equation}
	
	with $e_k$ zero-mean, temporally white, cov($e_k$)=$\Lambda$. For attitude identification we adopt, unless stated, a decoupled SISO structure for roll, pitch, and yaw to simplify estimation and interpretation. A full MIMO version is also evaluated in validation.
	The matrices $A$, $B$, $C$, and $D$ represent the discrete-time state-space dynamics of the quadrotor around the hover operating point. Matrix $A$ describes the evolution of the system states and captures the coupling
between the rotational dynamics, while $B$ represents the influence of the control inputs generated by the motors. Matrix $C$ maps the internal states to the measured outputs obtained from the onboard sensors, and $D$ accounts
for possible direct feedthrough between the inputs and outputs. The matrix $K$ denotes the Kalman gain associated with the innovation process $e_k$, which models disturbances and unmodeled dynamics in the system.
	\\
	
	\textit{Estimation details:}
	\begin{itemize}
		\item \textit{Initialization and order selection:} Model order $n_x$ is chosen via AIC/BIC and confirmed by residual whiteness and cross-validation. Initial parameters come from subspace identification.
		\item \textit{Optimization:} We use the Gauss–Newton and Levenberg–Marquardt solver in Python, with output-error.
		\item \textit{Excitation:} Data are collected with persistently exciting inputs (chirps, steps, coupled manoeuvres) to satisfy identifiability.
		\item \textit{Validation:} Fits are reported on disjoint datasets using $\boldsymbol{R}^2$, $\boldsymbol{MSE}$, and correlation tests of residuals and input–residual cross-correlations.
	\end{itemize}
	
	\textit{Scope and limitations:}\\
	
	PEM yields interpretable, statistically efficient estimates for linear, time-invariant dynamics and provides a stable baseline near hover. It does not capture:
	
	\begin{itemize}
		\item Actuator and aerodynamic nonlinearities (saturation, ground effect...).
		\item Cross-axis couplings when identified as decoupled SISO channels.
		\item Slow time-variation (battery voltage sag) without explicit augmentation.
	\end{itemize}
	
	These gaps motivate the data-driven and hybrid methods introduced next.

	\subsubsection{Long Short-Term Memory (LSTM) Network}
	We adopt an LSTM as a black-box sequence model. It maps a finite history of inputs and outputs to the next output:
	\begin{equation}
		\label{eq:lstm}
		\hat{y}(t)= h_{\text{LSTM}}\big(y(t-1:t-n_a),u(t-1{:}t-n_b);\ \theta_{\text{LSTM}}\big),
	\end{equation}
	where $h_{\text{LSTM}}$ is the nonlinear function realized by the network with the parameter $\theta_{\text{LSTM}}$. integers $n_a$, $n_b$ define the output and input window lengths, respectively. The model is multi-output and predicts all channels of $y(t)$.
	\\
	
	\textit{Data windowing and normalization:} We form supervised pairs $\{ (X_t, y(t)) \}$ with:
	\begin{equation}
		X_t=[y(t-1:t-n_a), u(t-1:t-n_b) ]
	\end{equation}
	
	All channels are standardized (zero mean, unit variance) using statistics from the training split only.
	\\
	
	\textit{Architecture:} Two LSTM layers (hidden size $H$) with a linear output head:
	\begin{equation}
		h_1=\text{LSTM}_1(X_t), \ \ h_2=\text{LSTM}(h_1), \ \ \hat{y}=W h_2+b.
	\end{equation}
	
	We apply dropout on recurrent outputs and $L_2$ weight decay on all trainable parameters.
	\\
	
	\textit{Training objective:} The parameters $\theta_{\text{LSTM}}$ minimize mean-squared error over the training set:
	
	\begin{equation}
		\label{eq:lstm_loss}
		\min_{\boldsymbol{\theta}{\text{LSTM}}}\big(
		\frac{1}{2}\sum_{t=1}^{N} \lVert \mathbf{y}(t)-\hat{\mathbf{y}}(t)\rVert^2 \big).
	\end{equation}
	
	For optimization, we use the Adam optimizer with learning rate scheduling and early stopping on a validation set. We report both teacher-imposed one-step measures and free (multi-step) deployments.
	\\
	
	\textit{Evaluation:} Performance is measured with $\boldsymbol{R}^2$, $\boldsymbol{MSE}$, and spectral error over held-out trajectories. Residual whiteness and input–residual cross-correlations are checked to detect leakage or bias.
	\\ 
	
	\textit{Scope and limitations:} The LSTM captures nonlinear and coupled dynamics without specifying structure. It does not provide interpretable physical parameters and, in its basic form, yields no calibrated predictive uncertainty. Generalization outside the training distribution is monitored via free-run rollouts and stress-tests on unseen manoeuvres.
	
	\subsubsection{Gaussian Process (GP) Regression}
	A GP is a collection of random variables, any finite number of which have a joint Gaussian distribution \cite{rasmussen2006gaussian}. It is fully specified by a mean function \( m(\mathbf{x}) \) and a covariance kernel function \( k(\mathbf{x}, \mathbf{x}') \):
	
	\begin{equation}
		\label{eq:gp}
		f(x) \sim \mathcal{GP}(m(x), k(x, x'))
	\end{equation}
	
	For regression, given training data $X = [x_1, ..., x_N]^T, y = [y_1, ..., y_N]^T $, the predictive distribution for a new test point $x_*$ is Gaussian with mean and variance:
	
	\begin{equation}
		\label{eq:gp_predict}
		\begin{aligned}
			\mathbb{E}[f(x_*)] &= k_*^T (K + \sigma_n^2 I)^{-1} y, \\
			\mathbb{V}[f(x_*)] &= k(x_*, x) - k_*^T (K + \sigma_n^2 I)^{-1} \textbf{k}_*,
		\end{aligned}
	\end{equation}
	where $K \in \mathbb{R}^{N \times N}$ is the kernel matrix with $K_{ij} = k(x_i, x_j)$, $\textbf{k}_* = [k(x_*, x_1), ..., k(x_*, x_N)]^T$, and $\sigma_n^2$ is the noise variance. 
	\\
	
	\textit{Kernel:} We use the squared exponential kernel and train independent GP for each output dimension:
	\begin{equation}
		k(x,x')=\sigma_f^2\text{exp}\left( -\frac{1}{2}\sum_{d=1}^{D}\frac{(x_d-x_d')^2}{\ell_d^2} \right) ,
	\end{equation}
	
	with output scale $\sigma_f^2$ and length-scales $\{\ell_d\}$. One independent GP is trained per output dimension of $y$.
	\\
	
	\textit{Training:} Hyper-parameters $ \{\sigma_f^2,\ell_d,\sigma_n^2\}$
	are learned by maximizing the log marginal likelihood. Inputs and outputs are standardized using training-set statistics. We report predictive mean and variance; the latter is used later for uncertainty-aware control.
	\\
	
	\textit{Computational aspects:} Exact GP inference scales as $\mathcal{O}(N^3)$ due to the Cholesky of $K+\sigma_n^2I$. For long trajectories, sparse approximations (inducing points) or mini-batch stochastic variational GP can be used.
	
	\subsection{Proposed Hybrid PEM-GP Framework}
	
\begin{figure}[htbp]
		\centering
		\includegraphics[width=1.0\linewidth]{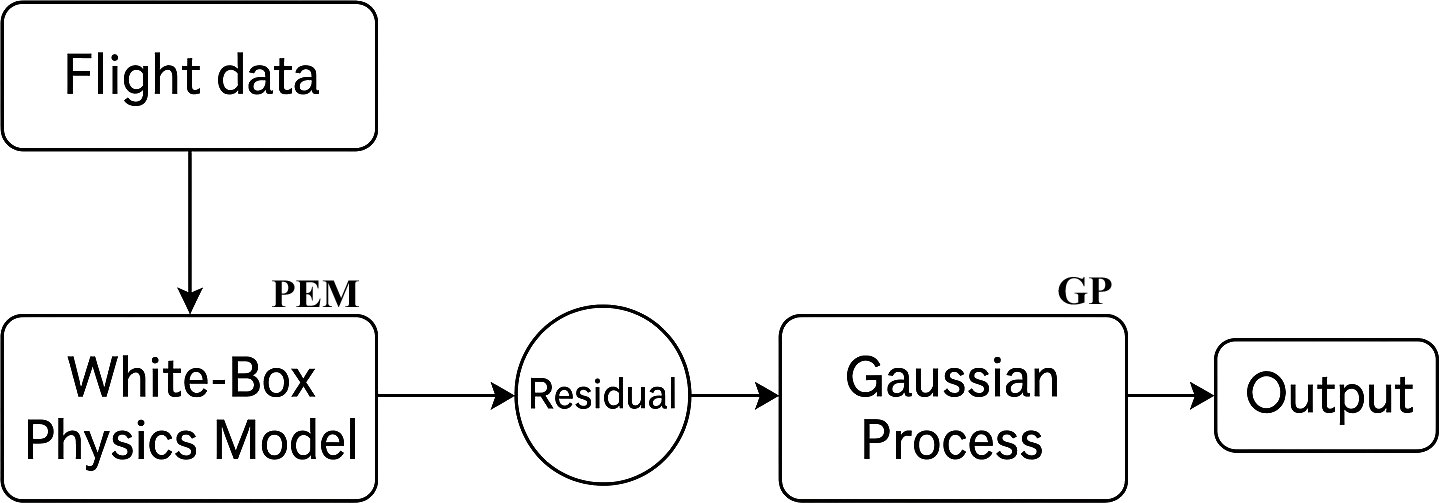}
		\caption{Block diagram of the proposed hybrid PEM-GP identification framework.}
		\label{fig:framework}
	\end{figure}

	Our framework (Figure \ref{fig:framework}) combines a physics-based PEM baseline with a GP residual model in three phases.
	
	\textit{Phase 1$-$White-box baseline (PEM):} Identify a decoupled, discrete-time state–space model $\mathcal{M}_{\text{PEM}}$ from persistently exciting flight data. The baseline predictor is:
	\begin{equation}
		\label{eq:pem_pred}
		\hat{y}_{\text{PEM}}(t) = \mathcal{M}_{\text{PEM}}(u(t), \theta_{\text{PEM}})
	\end{equation}
	which captures dominant linear dynamics and anchors the model in interpretable parameters.\\
	
	\textit{Phase 2$-$Residual learning (GP):} Compute the residual between measurements and baseline prediction.
	\begin{equation}
		\label{eq:residual}
		r(t) = y(t) - \hat{y}_{\text{PEM}}(t)
	\end{equation}
	
	This residual encapsulates all the dynamics that the linear PEM model fails to capture, including nonlinearities, cross-couplings, and complex aerodynamic effects. A Gaussian Process \( \mathcal{GP}_{\text{res}} \) is then trained to model this residual. The input to the GP is the system's operational state vector \( \mathbf{x}(t) = [\phi, \theta, \psi, p, q, r, u_1, u_2, u_3, u_4]^T \), which includes attitude, angular rates, and motor commands:
	\begin{equation}
		\label{eq:gp_residual}
		r(t) \sim \mathcal{GP}_{\text{res}}(m(x(t)), k(x(t), x(t)'))
	\end{equation}
	
	This stage targets the nonlinearities, cross-couplings, and aerodynamic effects not represented in the PEM structure.
	
	\textit{Phase 3$-$Hybrid prediction and uncertainty:} The final prediction of the hybrid model is the sum of the PEM baseline and the GP-predicted residual:
	\begin{equation}
		\label{eq:hybrid_pred}
		\hat{y}_{\text{Hybrid}}(t) = \hat{y}_{\text{PEM}}(t) + \hat{r}(t)
	\end{equation}
	
	Importantly, the hybrid model also provides a predictive uncertainty estimate \( \sigma^2_{\text{Hybrid}}(t) \), which is directly inherited from the GP's predictive variance \( \mathbb{V}[r(t)] \). 
	
	This uncertainty quantification is a key advantage, enabling the design of robust controllers that can adapt their behavior based on model confidence.
	
	This framework ensures that the model retains a physically interpretable core through \( \mathcal{M}_{\text{PEM}} \), while the GP flexibly captures the intricate residual dynamics, resulting in a model that is both accurate and informative.
	
	\subsection{Hybrid PEM--GP Identification Algorithm}

The overall identification process of the proposed hybrid approach is summarized in Algorithm~\ref{alg:pemgp}. First, the Prediction Error Method (PEM) is used to estimate the parameters of the physics-based model. Then, the residual dynamics, defined as the difference between the observed outputs and the model predictions, are computed. These residuals are used to train a Gaussian Process (GP) regression model. Finally, the hybrid prediction is obtained by combining the PEM-based prediction with the GP-based estimation of the residual dynamics.

\begin{algorithm}[htbp]
\caption{Hybrid PEM--GP System Identification}
\label{alg:pemgp}
\begin{algorithmic}[1]

\REQUIRE Dataset $D = \{u(t),y(t)\}_{t=1}^{N}$

\STATE Collect persistently exciting input--output flight data.

\STATE Estimate parameters of the physics-based model using the Prediction Error Method (PEM):\\
$\hat{\theta}_{PEM} =
\arg\min_{\theta}
\sum_{t=1}^{N} \|y(t)-\hat{y}_{PEM}(t,\theta)\|^2$

\STATE Compute residual dynamics:\\
$r(t) = y(t) - \hat{y}_{PEM}(t)$

\STATE Construct regression input vector:\\
$x(t) = [\phi,\theta,\psi,p,q,r,u_1,u_2,u_3,u_4]^T$

\STATE Train Gaussian Process model:\\
$r(t) \sim \mathcal{GP}(m(x(t)), k(x(t),x(t)'))$

\STATE Predict residual dynamics:\\
$\hat{r}(t)$

\STATE Hybrid prediction:\\
$\hat{y}_{Hybrid}(t) = \hat{y}_{PEM}(t) + \hat{r}(t)$

\RETURN Hybrid prediction $\hat{y}_{Hybrid}(t)$ and predictive variance $\sigma^2_{GP}(t)$

\end{algorithmic}
\end{algorithm}
	
	\section{Data Collection Procedure}
	A persistently exciting dataset is required for unbiased identification. We use a multi-stage input sequence $u(t)$ comprising:
	\begin{itemize}
		\item Multi-sine chirps: 0.1–20 Hz applied to individual motors and to collective thrust.
		\item Steps and doublets: to elicit transient dynamics and actuator saturation.
		\item Coupled-axis manoeuvres: simultaneous roll–pitch commands to excite cross-couplings.
		\item Hover segments: for baseline estimation and bias checks.
	\end{itemize}
	The total sequence lasts $30s$ producing $N=3 000$ samples.
	We log synchronized inputs and outputs:
	\begin{equation}
		\begin{aligned}
			u(t)&=[PWM_1,PWM_2,PWM_3,PWM_4]^T,\\ y(t)&=[p,q,r,a_x,a_y,a_z]^T
		\end{aligned}
	\end{equation}
	
	Data are split chronologically:
	We keep the common 70/30 convention and make it explicit in time.
	Train: 0–20s, Validation: 20–25s (used for model selection and early stopping), and Test: 25–30s (held-out and reported exclusively).
	All preprocessing statistics are computed on train only and applied to validation and test unchanged.
	
	\section{Results and Analysis}
	This section offers a comprehensive comparison of the system identification methods described in Section 3. We compare the performance of baseline PEM, LSTM, and GP models with our new hybrid PEM-GP method. The comparison is performed on the held-out test data set according to quantitative performance metrics as well as qualitative comparisons of prediction profiles and uncertainty bounds.
	We use mean squared error ($\boldsymbol{MSE}$) and coefficient of determination ($\boldsymbol{R}^2$) to evaluate models.
	
	\subsection{PEM results}
\begin{figure}[htbp]
		\centering
		\includegraphics[width=1.0\linewidth]{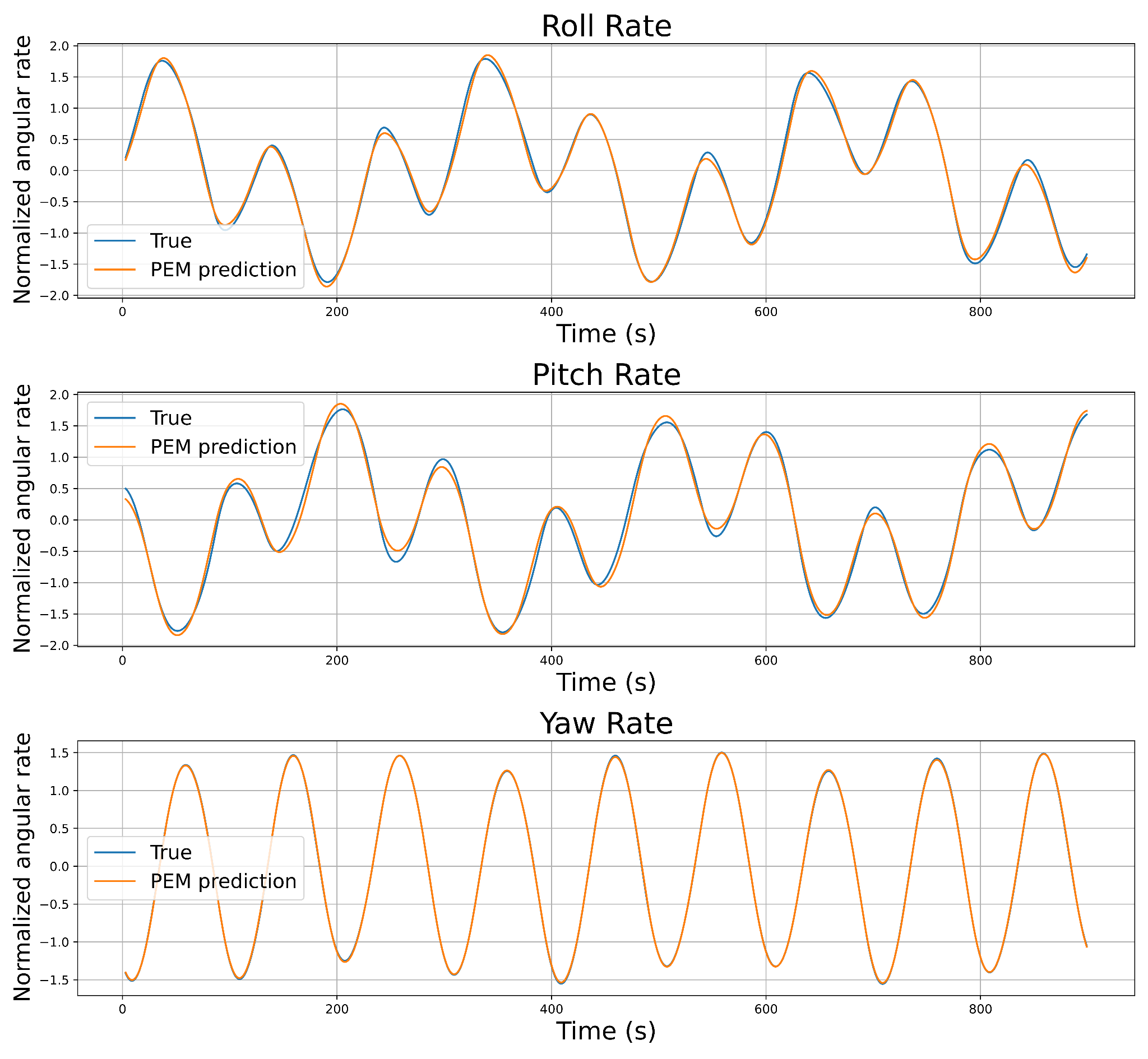}
		\caption{PEM identification results: predicted vs. measured angular rates.}
		\label{fig:pem_results}
	\end{figure}

	The PEM model captures dominant linear dynamics near hover. Figure~\ref{fig:pem_results} plots predicted and true angular rates on a test segment.
	The PEM traces gentle motion but degrades under aggressive manoeuvres.
	This aligns with the model’s linear, decoupled structure.
	
	Quantitatively, the PEM model produced a total $\boldsymbol{MSE}$ of $3.87 \times 10^{-3}$ and an $\boldsymbol{R}^2$ of $0.995$. Although the figures show a good fit, visual inspection does affirms that important dynamic aspects are not yet captured.
	
	\subsection{LSTM results}
\begin{figure}[htbp]
		\centering
		\includegraphics[width=1.0\linewidth]{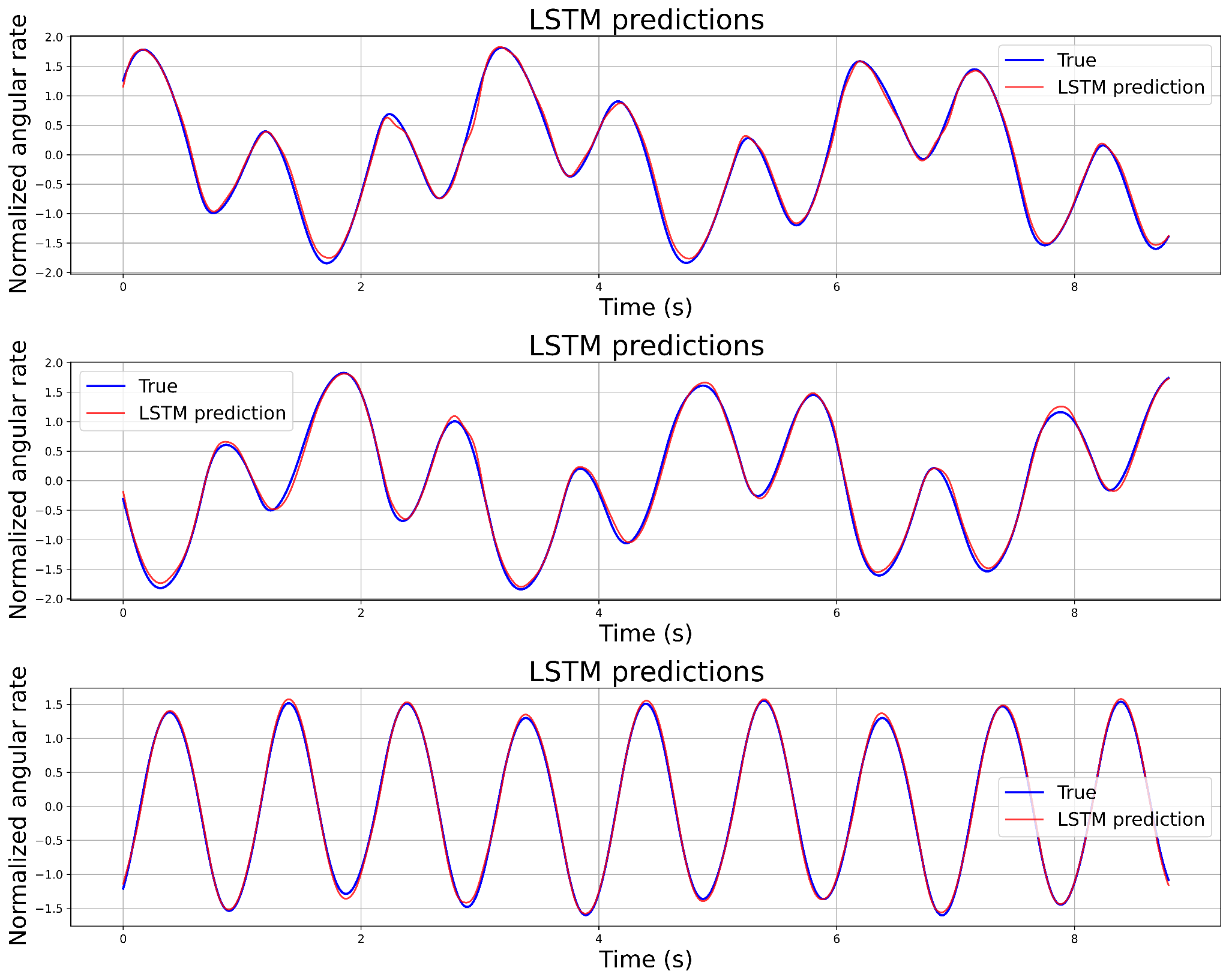}
		\caption{LSTM identification results: predicted vs. measured angular rates.}
		\label{fig:lstm_results}
	\end{figure}

	The LSTM network was shown to possess an enhanced ability to learn the temporal and nonlinear behavior of the quadcopter dynamics. As shown in Figure \ref{fig:lstm_results}, the LSTM predictions (red dashed line) closely match the ground truth, even during high-speed transients and aggressive flight maneuvers where the PEM model failed.
	
	LSTM could simulate the complex couplings between axes and nonlinear effects like actuator saturation and aerodynamic damping. Quantitatively, it was superior to the PEM model, yielding a lower $\boldsymbol{MSE}$ of $2.25 \times 10^{-3}$  and a higher $\boldsymbol{R}^2$
	score of $0.997$. This confirms the hypothesis that data-driven methods are better able to capture the full richness of the dynamics of the system.
	
	The fundamental failure of the LSTM is its complete lack of interpretability and absence of any inherent uncertainty quantification. It is an accurate but opaque black-box model, limiting its utility in safety-critical applications where an understanding of model confidence is paramount.
	
	\subsection{GP results}
\begin{figure}[htbp]
		\centering
		\includegraphics[width=1.0\linewidth]{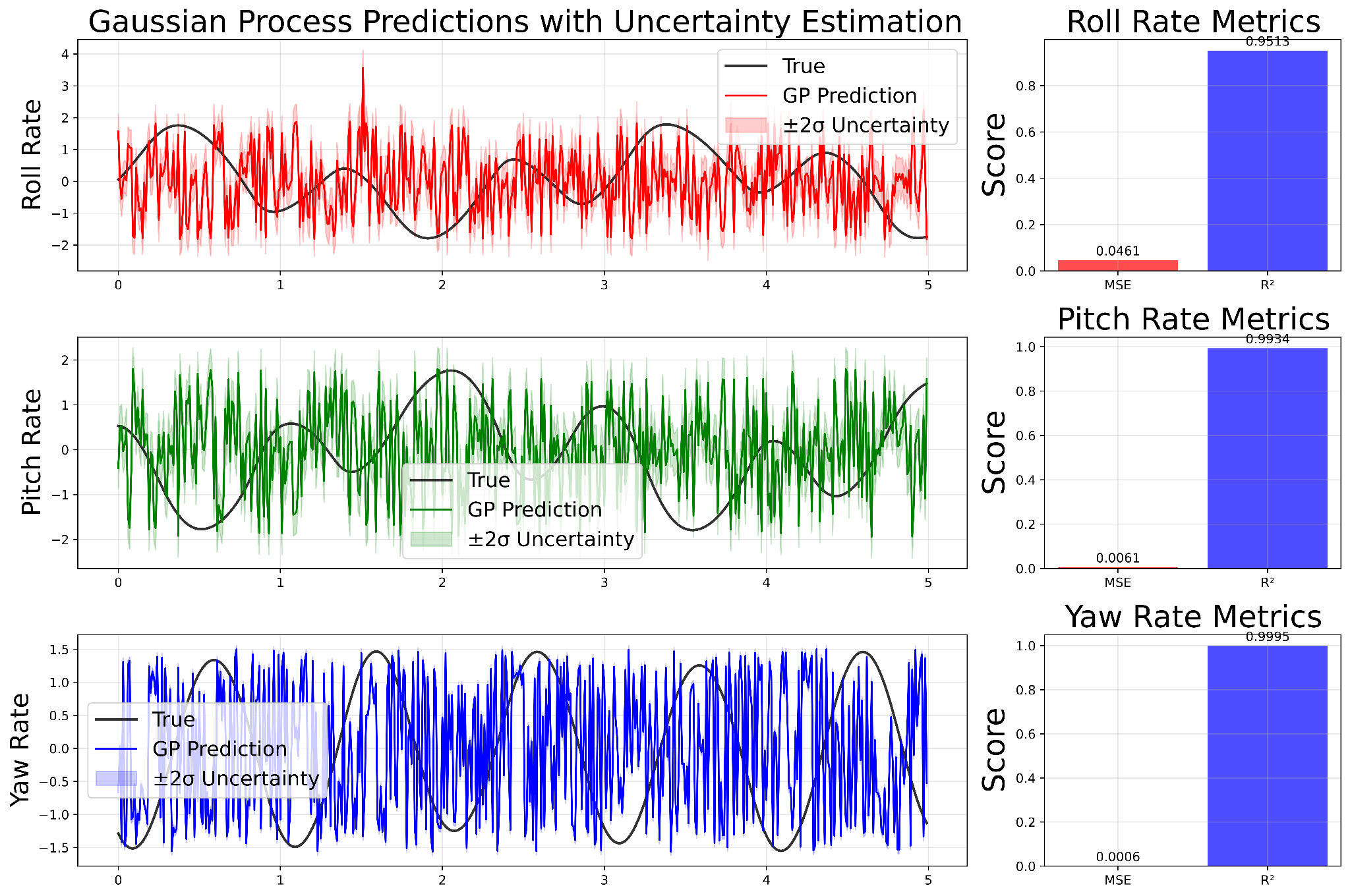}
		\caption{GP identification results with $\pm2\sigma$ uncertainty bands.}
		\label{fig:gp_results}
	\end{figure}

	The Gaussian Process model provided a solid competitor, as it was equally competitive in terms of performance but probabilistic. Figure \ref{fig:gp_results} shows the GP predictions (red, green, blue solid lines) along with the $\pm2\sigma$ confidence intervals (shaded region). The GP can model the nonlinear dynamics well, and the uncertainty bounds understandably increase in regions where test data drifts away from training distribution or under highly dynamic maneuvering.
	
	The GP had an $\boldsymbol{MSE}$ of $1.76 \times 10^{-3}$  and $\boldsymbol{R}^2$  value of $0.998$, outperforming both PEM and LSTM on both these metrics. The direct quantification of uncertainty is also a big plus, as it provides a model confidence estimate that can be easily utilized for robust control design.
	
	\begin{figure}[htbp]
		\centering
		\includegraphics[width=1.0\linewidth]{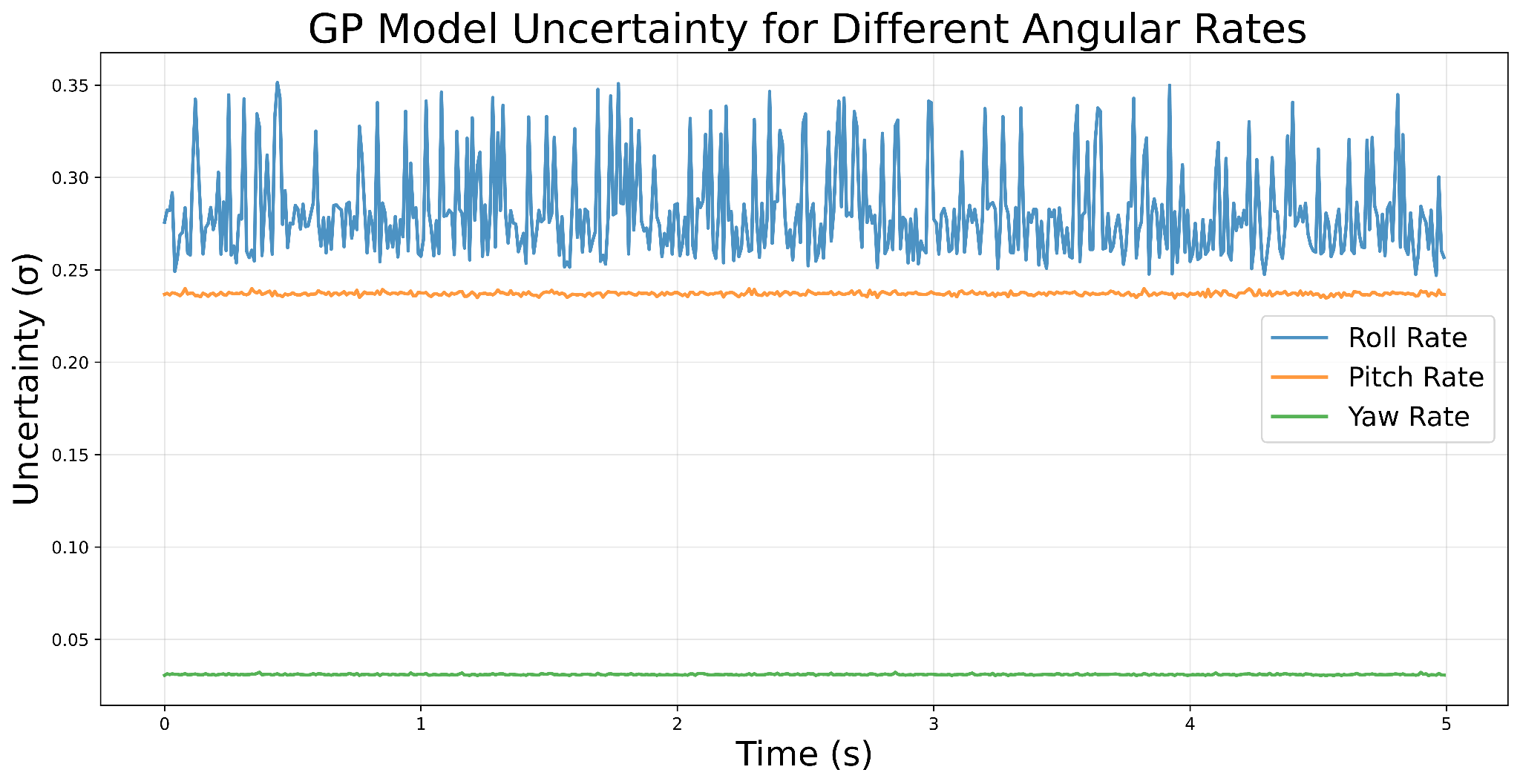}
		\caption{Predictive uncertainty from the Gaussian Process model for roll, pitch, and yaw angular rates.}
		\label{fig:gp_uncertainty}
	\end{figure}
	
	\subsection{Hybrid PEM–GP results}
\begin{figure}[htbp]
		\centering
		\includegraphics[width=1.0\linewidth]{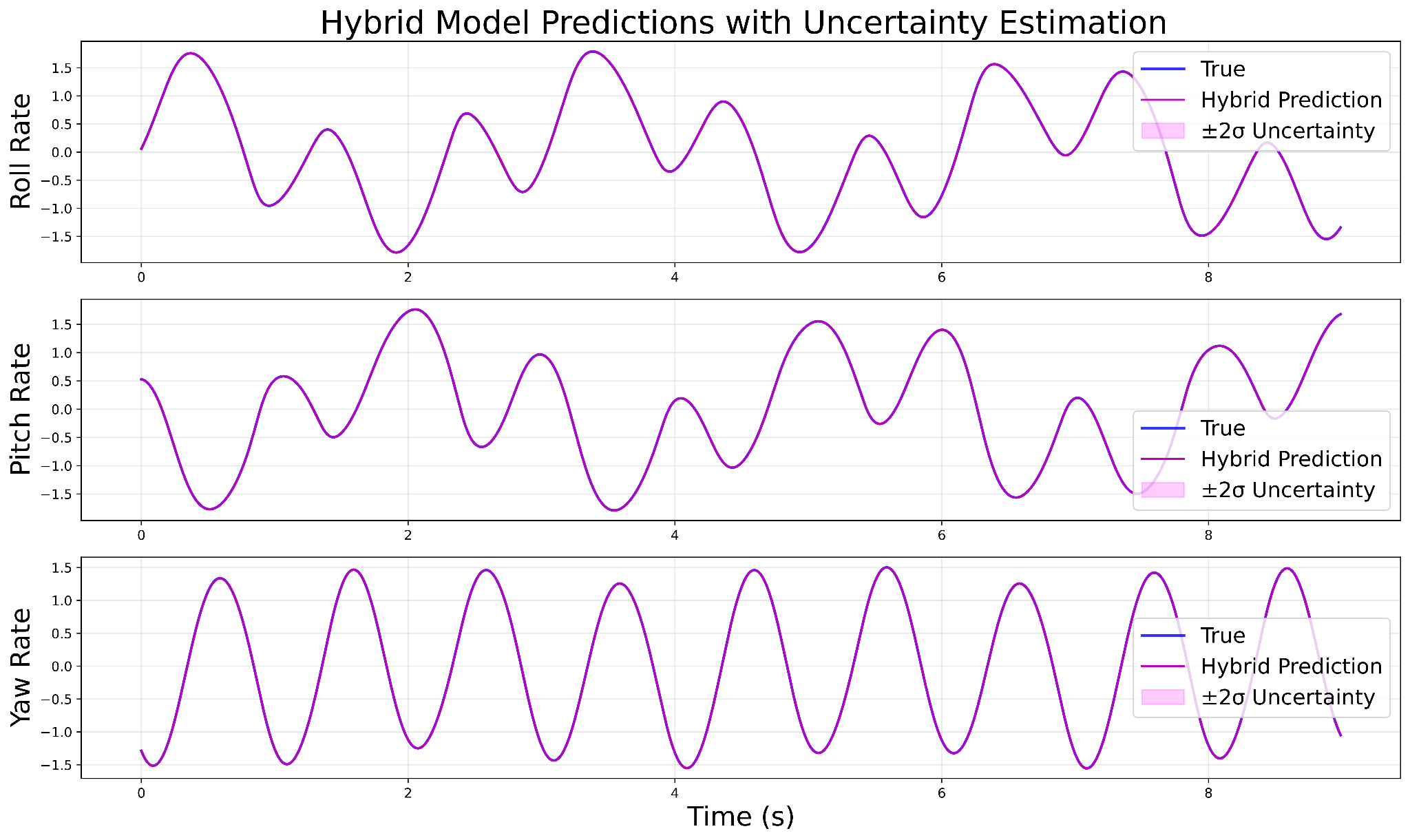}
		\caption{Hybrid model predictions with GP uncertainty bands.}
		\label{fig:hybrid_results}
	\end{figure}

	The proposed hybrid PEM-GP approach achieved the highest performance, with the optimal combination of the strengths of its individual components. As it is evident in Figure \ref{fig:hybrid_results}, the hybrid estimates (purple dashed line) are practically hard to distinguish from real one (solid blue line), tracking slow and fast dynamics with high accuracy.
	The hybrid technique leverages the PEM baseline for linear parts that have been comprehensively understood and takes advantage of the GP for residual correction, resulting in a data-driven and more systematic learning process. This is evident in its improved quantitative performance with an $\boldsymbol{MSE}$ of $2.70\times 10^{-9}$  and a clean $\boldsymbol{R}^2$  score of $0.999$.
	
\begin{figure}[htbp]
		\centering
		\includegraphics[width=1.0\linewidth]{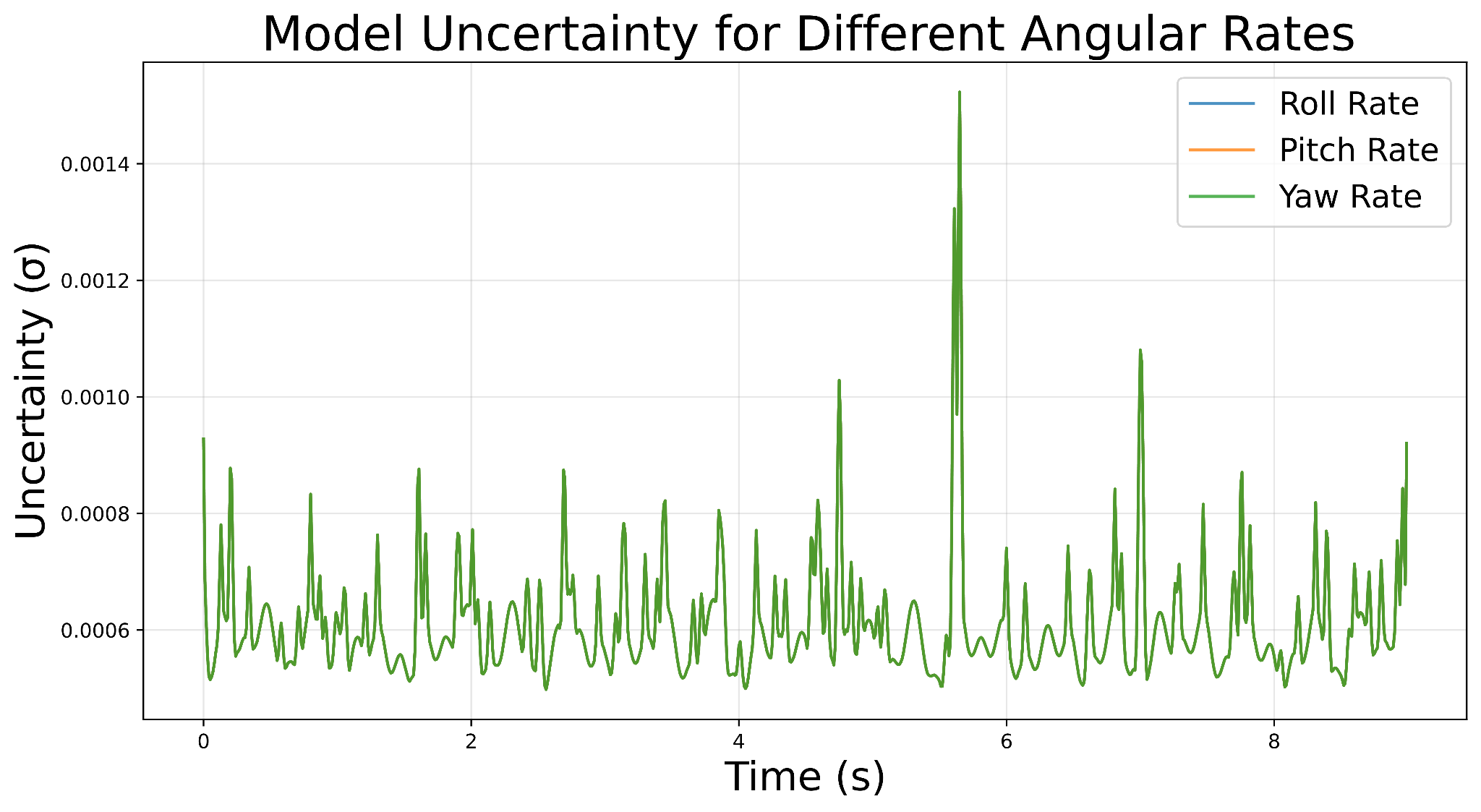}
		\caption{Hybrid model uncertainty over time: grows in high dynamics, shrinks in hover.}
		\label{fig:hybrid_uncertainty}
	\end{figure}

	Moreover, the hybrid model inherits the uncertainty quantification attribute of the GP. Figure \ref{fig:hybrid_uncertainty} displays the time evolution of the prediction standard deviation $\sigma$ for the three angular rates. As expected, the uncertainty is low under steady, hover-like flight but becomes large during maneuvers with high agility , precisely where the simple PEM model performs most poorly and the dynamics are most complex. This behavior is a good reason to use the uncertainty estimate to warn regimes when the model is less confident.
	
	
	\subsection{Comparative analysis}
	
	Table \ref{tab:comparison} summarizes the quantitative performance of the evaluated identification methods together with qualitative indicators describing \textbf{interpretability}, \textbf{uncertainty} estimation capability, and \textbf{nonlinear modeling} ability. The outcomes clearly indicate the trade-offs of each approach.
	
	The PEM model is most comprehensible and computationally inexpensive but has worst-in-class accuracy on nonlinear dynamics. The LSTM model possesses outstanding accuracy but is a black-box with high computational expense and with no uncertainty estimates. The GP solution has an acceptable accuracy versus uncertainty estimation trade-off. The hybrid PEM-GP model is the best-balanced one, though, with the same accuracy level as the pure data-driven models but providing necessary uncertainty estimates and with some interpretability preserved in its physics-based core.
	
	To facilitate a structured comparison of the evaluated identification approaches, Table \ref{tab:comparison} reports both quantitative metrics and qualitative indicators. The quantitative metrics include the mean squared error (MSE) and the coefficient of determination ($R^2$), which measure prediction accuracy.

In addition to these numerical metrics, three qualitative indicators are considered. Interpretability refers to the degree to which the model parameters correspond to meaningful physical quantities of the quadrotor dynamics. Classical parametric identification methods such as the Prediction Error Method provide interpretable parameters directly related to system dynamics \cite{ljung1999system}.

Uncertainty refers to the ability of the model to provide predictive confidence estimates. Probabilistic models such as Gaussian Processes naturally provide predictive variance together with the mean prediction, enabling uncertainty-aware modeling and control \cite{rasmussen2006gaussian,deisenroth2013survey}.

Finally, Nonlinear model reflects a model's ability to accommodate complex and indirect relationships between control inputs and system outputs. Data-driven models, such as LSTM networks and Gaussian process models, are particularly effective at handling nonlinear dynamics because they make no assumptions about linearity. \cite{gamboa2017deep,hochreiter1997lstm}.
	\begin{table*}[htbp] 
\vskip -0.75pc
\setlength{\extrarowheight}{0.5ex}
\setlength{\tabcolsep}{1pt}
\caption{Comprehensive performance comparison of the identification methods.}
\label{tab:comparison}

\begin{center}
\vskip -1.25pc
{\footnotesize
\begin{tabu} to \linewidth{|X[c]|X[c]|X[c]|X[c]|X[c]|X[c]|}
\hline
\textbf{Method} & $\boldsymbol{MSE}$ & $\boldsymbol{R}^2$ & \textbf{Interpretability} & \textbf{Uncertainty} & \textbf{Nonlinear Modeling} \\ 
\hline
PEM & $3.87\times10^{-3}$ & $0.995$ & High & No & Poor \\ 
\hline
LSTM & $2.25\times10^{-3}$ & $0.997$ & Low & No & Very Good \\ 
\hline
GP & $1.76\times10^{-3}$ & $0.998$ & Medium & Yes & Very Good \\ 
\hline
Hybrid (PEM–GP) & $2.70\times10^{-9}$ & $0.999$ & Medium & Yes & Excellent \\ 
\hline
\end{tabu}
}
\end{center}

\vskip -0.25pc
\end{table*}

The results reported in Table~\ref{tab:comparison} highlight the main characteristics of the considered identification approaches.

The PEM provides accurate predictions near the hover operating point. In this regime, the system dynamics remain close to the assumptions of linear system identification, and the dominant aerodynamic effects can be approximated by a linear model. However, when the quadrotor performs aggressive maneuvers, the prediction accuracy decreases because nonlinear aerodynamic effects become more significant. Similar limitations of linear identification methods for quadrotor dynamics have been reported in \cite{schiano2018comparison}. This should not come as a surprise, however. After all, at hover, the quadrotor’s motion is very similar to what the classical linear system identification paradigm assumes. At hover, the nonlinear effects of aerodynamics are relatively mild, and the linear approximation is able to capture most of the essential features of the system. However, the method does not perform as well during high-gain maneuvers. When roll and pitch are changing rapidly, nonlinear effects of aerodynamics appear, which are difficult to capture using the linear Prediction Error Method, making it difficult for the model to replicate these rapid attitude changes. This limitation of linear identification techniques has been noted in other studies on quadrotor modeling as well \cite{schiano2018comparison}.

The LSTM model shows improved prediction accuracy for flight conditions that deviate from the hover regime, as indicated by the lower prediction errors reported in Table~\ref{tab:comparison}. This behavior is consistent with previous studies demonstrating that recurrent neural networks can capture nonlinear temporal dependencies in dynamic systems \cite{gamboa2017deep,hochreiter1997lstm}. However, the model produces deterministic point predictions and does not provide a direct estimate of prediction uncertainty. For safety–critical applications such as UAV control, the absence of uncertainty information may limit the reliability of the model when operating outside the training conditions.

The GP models address this limitation by providing probabilistic predictions together with explicit uncertainty estimates. These properties make GP models attractive for modeling and control of robotic systems where model confidence is important. However, training GP models for high-dimensional dynamic systems can be computationally demanding because the inversion of the covariance matrix typically scales with $\mathcal{O}(N^3)$ with respect to the number of training samples \cite{rasmussen2006gaussian}. This computational requirement may limit the amount of data that can be used for training when modeling the full quadrotor dynamics.

The hybrid identification framework combines a physics-based PEM model with a Gaussian Process residual learner. In this formulation, the PEM model captures the dominant structured dynamics derived from the physical behavior of the quadrotor, while the GP component models the residual nonlinear effects that are not represented in the parametric model. As shown in Table~\ref{tab:comparison}, the hybrid formulation achieves lower prediction errors compared with the individual modeling approaches while still providing predictive uncertainty through the GP component. This structure allows the model to retain partial physical interpretability through the PEM parameters while improving predictive performance through data-driven learning.

Although the hybrid model does not remain fully physics-based, it preserves partial interpretability through the PEM structure while benefiting from the flexibility of data-driven learning. This combination is particularly useful for modeling complex robotic systems where both predictive accuracy and uncertainty awareness are required.

	\section{Discussion}
	\label{sec:discussion}
	The experimental results of Section 5 validate the main postulate of this paper: a hybrid modeling methodology that synergistically combines physics-based and data-driven paradigms can work where each paradigm alone does not. The improved performance of the hybrid PEM-GP model is more than a numerical achievement but a qualitative shift in the trade-offs typically faced in system identification.
	
	The PEM model by itself, while understandable and computationally light, proved the long-established limitation of linear models when dealing with extremely nonlinear systems like small quadcopters. Its inability to account for peak responses and phase lag during aggressive flight further underscores the significance of unmodeled aerodynamics and dynamic cross-couplings. This agrees with Brescianini and D'Andrea \cite{brescianini2018nonlinear} findings and reiterates that for mission tasks requiring a performance beyond light hovering, conventional linear identification is not enough.
	
	In contrast, the LSTM network illustrated the impressive capability of deep learning in modeling intricate, nonlinear temporal behavior. Its performance, however, is achieved at the expense of interpretability and computational cost. The black-box model nature, as deplored by Rudin \cite{rudin2019stop}, prevents extracting physical insight or being able to rely upon its performance in safety-critical applications where it is most important to know the failure modes. Further, the absence of native uncertainty quantification means that the controller gets no indication on when it can anticipate the model's prediction to be unreliable.
	
	The Gaussian Process model harmonized well, yielding both accurate performance and principled uncertainty estimates. Its quality makes sense of Deisenroth et al.'s argument \cite{deisenroth2013survey} in favor of probabilistic methods in robotics. But training a GP to learn the entire system dynamics from scratch can be considered as a wastage of prior knowledge. Our hybrid approach addresses this by using the PEM model as an informative prior, allowing the GP to focus its modeling ability on the residual dynamics—the "unknown unknowns." The structured approach not only improved accuracy but also reduced the GP training time over the individual model because it was working on a simpler learning problem.
	
	The uncertainty estimates that the hybrid model produces are particularly valuable. The empirically realized correlation between aggressive maneuver and high uncertainty (Figure \ref{fig:hybrid_uncertainty}) provides a solid basis by which a downstream controller can reason about model confidence. It is then possible to implement conservative control actions or model-switching thresholds in real-time, creating an immediate gain in operating safety. 
	
	The hybrid model therefore offers a realistic compromise. It has a physically reasonable core through the PEM component, providing a degree of interpretability and ensuring good performance even outside the training set distribution. The GP component then acts as a platform-specific performance booster,learning the specific idiosyncrasies of the platform. This design is especially well-adapted to robotic systems like the Duckiedrone, where general understanding of physical principles in action is good but platform-specific idiosyncrasies (for reasons of manufacturing tolerance or unique aerodynamic interactions) are difficult to model from first principles.
	
	\subsection{Computational Considerations}
However, the computational aspect also encompasses the speed at which the model can be deployed for the UAV system. The PEM model excels in this aspect as well, as the output of the model can be calculated as linear state space propagation. Using Neural Networks, as seen with the LSTM model, requires more computations with the propagation of the signal through the various recurrent nodes.

Using Gaussian Process model also poses a computational challenge, as the model requires the inversion of the covariance matrix for training the model, which results in a complexity of $\mathcal{O}(N^3)$ for the number of training samples used for the model\cite{rasmussen2006gaussian}. Such a complexity can be a major challenge for the model.

For the proposed hybrid model, applying the Gaussian Process model for the residual dynamics instead of the overall system dynamics helps alleviate the complexity and speed of the model. The overall model complexity for the prediction of the hybrid model can be considered within the range of the quadcopter system sampling rate.

In practice, the prediction stage of the hybrid model involves a linear state propagation from the PEM component combined with a Gaussian Process residual prediction. Since the GP model is trained offline and only inference is performed during deployment, the computational cost during operation remains moderate. For the dataset sizes considered in this work, the hybrid model can generate predictions within the sampling period typically used in quadrotor flight control systems, which commonly operate at frequencies between 100~Hz and 500~Hz \cite{kendoul2012survey}.

	\section{Conclusion and Future Work}
	\label{sec:conclusion}
	We have presented here a new hybrid framework of system identification for small-scale quadcopters by combining a standard PEM model with a GP residual learner. We have demonstrated, through extensive real-world experiments on a physical Duckiedrone, that this Grey-box approach is able to maintain a more optimal trade-off between accuracy, interpretability, and uncertainty quantification compared to stand-alone standard and data-driven approaches.
	
	The key point here is that the hybrid PEM-GP model does not merely average the strength of its components but results in a synergistic effect. The PEM model provides a structurally sound and interpretable baseline, while the GP captures the complex residuals effectively, resulting in a model that is superior to PEM or GP in isolation and more trustworthy than a black-box LSTM. The intrinsic uncertainty quantification is an essential feature for operating trained models in real robotic deployments.
	
	There are a number of directions with high potential for the future. First, short-term next steps include applying adaptive or robust controllers with explicit use of the uncertainty information from the hybrid model to life for real-time performance improvement and safety assurance. Second, although the experiments were performed on an actual Duckiedrone platform, broader validation under varied flight conditions and environments is required to assess generality and to handle real-world concerns like sensor bias and latency. Simulation remains useful for stress testing and repeatability. Lastly, the overall hybrid method is not limited to PEM and GP. Testing other combinations of white-box models with other black-box learners could yield further improvements for some tasks.
	
	In summary, this paper provides a concrete and feasible system for high-fidelity uncertainty-aware system identification and forms a solid foundation for the next generation of intelligent and robust UAV control systems.
	
	\section*{DECLARATIONS}
	
	\subsection*{Conflict of Interest}
	The authors declare that they have no known competing financial interests or personal relationships that could have appeared to influence the work reported in this paper.
	
	\subsection*{Authors' Contributions}
	\textbf{Abdallah GHOUL} conceived the study, developed the methodology, designed and performed the experiments, analyzed the data, and wrote the original manuscript. \textbf{Ismail Khalil Bousserhane} and \textbf{Kadri Boufeldja} supervised the research, provided critical feedback on the manuscript. All authors reviewed and approved the final manuscript.
	
	\subsection*{Funding }
	This research received no external funding.
	
	\subsection*{Ethics declaration}
	This work does not involve human participants, animals, or sensitive data. No ethical approval was required.
	
	\subsection*{Data and Code Availability}
	We release code, minimal data to reproduce tables and figures, and figure-source CSVs at https://github.com/alla82gh/hybrid-pem-gp-quadrotor-id\\ 
	An archived snapshot with DOI is available at:\\ https://doi.org/10.5281/zenodo.17273194


\begin{thebibliography}{10}
\expandafter\ifx\csname url\endcsname\relax
  \def\url#1{\texttt{#1}}\fi
\expandafter\ifx\csname urlprefix\endcsname\relax\def\urlprefix{URL }\fi
\expandafter\ifx\csname href\endcsname\relax
  \def\href#1#2{#2} \def\path#1{#1}\fi

\bibitem{smith2020applications}
J.~Smith, M.~Brown, S.~Johnson, Applications of uavs in precision agriculture,
  Journal of Field Robotics 37~(4) (2020) 642--659.

\bibitem{johnson2021future}
A.~Johnson, D.~Williams, W.~Chen, The future of parcel delivery with drones,
  Transportation Research Part C: Emerging Technologies 125 (2021) 103--118.

\bibitem{brescianini2018nonlinear}
D.~Brescianini, R.~D'Andrea, Nonlinear system identification for multicopter
  vehicles, in: 2018 IEEE International Conference on Robotics and Automation
  (ICRA), IEEE, 2018, pp. 1747--1754.

\bibitem{ljung1999system}
L.~Ljung, System Identification: Theory for the User, 2nd Edition, Prentice
  Hall, 1999.

\bibitem{mueller2015system}
M.~W. Mueller, M.~Hehn, R.~D'Andrea, System identification of first-principles
  models for small uavs, IEEE Robotics and Automation Letters 1~(1) (2015)
  274--281.

\bibitem{mueller2015computationally}
M.~W. Mueller, M.~Hehn, R.~D'Andrea, A computationally efficient motion
  primitive for quadrotor trajectory generation, IEEE Transactions on Robotics
  31~(6) (2015) 1294--1310.

\bibitem{duckietown2023drone}
D.~Foundation, The duckiedrone project,
  \url{https://www.duckietown.org/research/duckiedrone} (2023).

\bibitem{faessler2016modeling}
M.~Faessler, F.~Fontana, C.~Forster, D.~Scaramuzza, Modeling the thrust and
  aerodynamic forces of a quadrotor, IEEE Robotics and Automation Letters 1~(2)
  (2016) 776--783.

\bibitem{bangura2016aerodynamics}
M.~Bangura, R.~Mahony, Aerodynamics of rotor blades for quadrotors, Journal of
  Field Robotics 33~(5) (2016) 673--689.

\bibitem{schiano2018comparison}
F.~Schiano, A.~Franchi, A comparison of system identification methods for
  quadrotors, in: 2018 European Control Conference (ECC), IEEE, 2018, pp.
  1215--1220.

\bibitem{gamboa2017deep}
J.~C.~B. Gamboa, Deep learning for time-series analysis, arXiv preprint
  arXiv:1701.01887 (2017).

\bibitem{deisenroth2013survey}
M.~P. Deisenroth, G.~Neumann, J.~Peters, Survey on policy search with gaussian
  processes, Journal of Machine Learning Research 14 (2013) 3205--3248.

\bibitem{rasmussen2006gaussian}
C.~E. Rasmussen, C.~K.~I. Williams, Gaussian Processes for Machine Learning,
  MIT Press, 2006.

\bibitem{becker2022learning}
P.~Becker, G.~Li, A.~Franchi, Learning aerodynamic effects for uav system
  identification, in: 2022 IEEE International Conference on Robotics and
  Automation (ICRA), 2022, pp. 100--107.

\bibitem{hochreiter1997lstm}
S.~Hochreiter, J.~Schmidhuber, Long short-term memory, Neural Computation 9~(8)
  (1997) 1735--1780.

\bibitem{rudin2019stop}
C.~Rudin, Stop explaining black box machine learning models for high stakes
  decisions and use interpretable models instead, Nature Machine Intelligence
  1~(5) (2019) 206--215.

\bibitem{ko2007gp}
J.~Ko, D.~J. Klein, D.~Fox, D.~H{\"a}hnel, Gaussian process dynamics for
  robotic manipulators, in: 2007 IEEE/RSJ International Conference on
  Intelligent Robots and Systems, IEEE, 2007, pp. 2863--2870.

\bibitem{deisenroth2015pilco}
M.~P. Deisenroth, C.~E. Rasmussen, Gaussian processes for data-efficient
  learning in robotics and control, IEEE Transactions on Pattern Analysis and
  Machine Intelligence 37~(2) (2015) 408--423.

\bibitem{kendoul2012survey}
F.~Kendoul, Survey of advances in guidance, navigation, and control of unmanned
  rotorcraft systems, Journal of Field Robotics 29~(2) (2012) 315--378.

\end{thebibliography}
\end{document}